\pdfoutput=1
\documentclass[11pt]{article}

\usepackage[]{acl}

\usepackage{times}
\usepackage{latexsym}

\usepackage[T1]{fontenc}
\usepackage[utf8]{inputenc}

\usepackage{microtype}

\usepackage{booktabs}
\usepackage{array}
\usepackage{multirow}
\usepackage{rotating}
\usepackage{arydshln}
\usepackage{amssymb}
\usepackage{pifont}% http://ctan.org/pkg/pifont
\newcommand{\cmark}{\ding{51}}%
\newcommand{\xmark}{\ding{55}}%
\usepackage{algorithm}
\usepackage{algorithmic}
\usepackage{comment}
\usepackage{paralist}
\usepackage{fixltx2e}

\newcommand{\cwhw}{comment\textsubscript{w/ hateword}}
\title{Hate Speech and Counter Speech Detection:\\Conversational Context Does Matter}

\author{
  Xinchen Yu \\
  University of North Texas \\
  \small\texttt{xinchenyu@my.unt.edu} \\\And
  Eduardo Blanco \\
  Arizona State University \\
  \small\texttt{eduardo.blanco@asu.edu} \\\And
  Lingzi Hong \\
  University of North Texas \\
  \small\texttt{lingzi.hong@unt.edu} \\
}

\begin{document}
\maketitle
\begin{abstract}
Hate speech is plaguing the cyberspace along with
%the massive rise of
user-generated content.
%Adding counter speech has become an effective way to combat hate speech online. 
%Existing datasets and models target either
%(a) hate speech 
%or
%(b) hate and counter speech but disregard the context.
This paper investigates the role of conversational context in the annotation and detection of online hate and counter speech,
where context is defined as the preceding comment in a conversation thread.
We created a context-aware dataset for a 3-way classification task on Reddit comments: hate speech, counter speech, or neutral.
Our analyses indicate that context is critical to identify hate and counter speech:
human judgments change for most comments depending on whether we show annotators the context.
A linguistic analysis draws insights into the language people use to express hate and counter speech. 
Experimental results show that neural networks obtain significantly better results if context is taken into account.
We also present qualitative error analyses shedding light into
(a) when and why context is beneficial
and
(b) the remaining errors made by our best model when context is taken into account.
\end{abstract}

\section{Introduction}
\label{s:introduction}
The advent of social media has democratized public discourse on an unparalleled scale. 
Meanwhile, it is considered a particularly conducive arena for hate speech \cite{caiani2021online}. 
%We use the term \textit{hate speech} as an umbrella term to refer to \textit{hate}, \textit{abusive} \cite{nobata2016abusive}, \textit{offensive} \cite{hateoffensive}, and \textit{cyberbullying} \cite{van-hee-etal-2015-detection} language. 
Online hate speech is prevalent and can lead to serious consequences. 
At the individual level, the victims targeted by hate speech are frightened by online threats that may materialize in the real world \cite{olteanu2018effect}. 
At the societal level, it has been reported that there is an upsurge in offline hate crimes targeting minorities \cite{olteanu2018effect,farrell2019exploring}. 

\begin{table}[t]
	\small
	\centering
	
	\begin{tabular}{p{0.6cm} p{6.1cm}}
		%\multicolumn{2}{c}{TARGET LABEL: NEUTRAL $\Rightarrow$ HATE} \\
		\toprule
		\emph{Parent} & As an average height male, idgaf how tall you are, if that's your issue then spend the money and get a better seat, or just f**king make the seat selection online to get more space.  \\
		\emph{Target} & Found the short guy! \\ \addlinespace
		
		\multicolumn{2}{l}{~-\emph{Target} is Neutral if considering only \emph{Target}.} \\
		\multicolumn{2}{l}{~-\emph{Target} is Hate if considering \emph{Parent} and \emph{Target}.} \\
		\midrule
		%\multicolumn{2}{c}{} \\
		%\multicolumn{2}{c}{TARGET LABEL: HATE $\Rightarrow$ COUNTER-HATE} \\
		%\hline
		\emph{Parent}  &  I deal with women all day with my job and this is how they are - extremely stupid, hate-filled, bizarre and they appreciate nothing.  \\
		%    Target & Maybe you're just an asshole if they treat you like that? \\ \addlinespace
		\emph{Target}& Maybe you're an a**hole if they treat you like that? \\ \addlinespace
		\multicolumn{2}{l}{~-\emph{Target}  is Hate if considering only \emph{Target}.} \\
		\multicolumn{2}{l}{~-\emph{Target} is Counter-hate if considering \emph{Parent} and \emph{Target}.} \\ \bottomrule
	\end{tabular}
\caption{Reddit comments (\emph{Target}s) deemed to be Hate, Neutral, or Counter-hate depending on whether one takes into account the previous comment (\emph{Parent}).
	%where \textit{Parent} affects \textit{Target} label. 
	%	In the top example, the label of \textit{Target} changed from Neutral to Hate after providing annotators with \textit{Parent}.
	%	 In the bottom example, \textit{Target} label changed from Hate to Counter-hate.
}
\label{t:problem-examples}
\end{table}

There are two common strategies to combat online hate: 
disruption 
and 
counter speech. 
Disruption refers to blocking hateful content or users. % temporally or permanently on a platform. 
To scale this strategy, researchers have proposed methods
%automated detection algorithms have been invented
to identify hate~\cite{waseem-hovy-2016-hateful,hateoffensive,nobata2016abusive}. 
While these interventions could de-escalate the impact of hate speech, they may violate online free speech \cite{DBLP:conf/icwsm/MathewSTRSMG019}. 
Additionally, attacks at the micro-level may be ineffective as hate networks often have rapid rewiring and self-repairing mechanisms~\cite{johnson2019hidden}. 
Counter speech refers to the ``direct response that counters hate speech''~\cite{DBLP:conf/icwsm/MathewSTRSMG019}.
%, which is considered a remedy to address hate speech problems. 
It has been shown to be more effective in the long term than disruption in theoretical and empirical studies~\cite{richards2000counterspeech,mathew2020hate}. 
Identifying hate and counter speech in natural conversations is critical to 
understand effective counter speech strategies 
and the generation of counter speech. % against hate speech. 

Most corpora with either hate speech (Hate) or counter speech (Counter-hate) annotations do not include conversational context.
Indeed, they annotate a user-generated comment as Hate or Counter-hate based on the comment in isolation~\cite{hateoffensive,waseem-hovy-2016-hateful,DBLP:conf/icwsm/MathewSTRSMG019,he2021racism}. 
Therefore, systems trained on these corpora fail to consider the effect of contextual information on the identification of Hate and Counter-hate. 
Recent studies have shown that context affects annotations in toxicity and abuse detection~\cite{pavlopoulos-etal-2020-toxicity, DBLP:journals/corr/abs-2103-14916}. 
We further investigate the effect of context on the task of identifying Hate and Counter-hate.
Table~\ref{t:problem-examples} shows examples where a comment, denoted as \emph{Target}, is Hate, Neutral or Counter-hate depending on whether the preceding comment, denoted as \emph{Parent}, is taken into account.\footnote{The examples in this paper contain hateful content. We cannot avoid it due to the nature of our work.}
%differently when the previous comment (\textit{Parent}) is provided or not. 
In the top example, the \emph{Target} goes from Neutral to Hate when taking into account the \emph{Parent}: 
it becomes clear that the author is disparaging short men.
% when the annotators are able to see \textit{Parent}. 
In the bottom example, the \emph{Target} goes from Hate to Counter-hate as the author uses offensive language to counter the hateful content in the \emph{Parent}. 
%\textit{Target} alone is labeled as Hate in the bottom example as it uses a swear word to describe a person. 
%However, the comment turns into Counter-hate when \textit{Parent} is shown. 
%\textit{Target} tries to debunk hate speech using the hostile language, which is a commonly seen counter hate strategy
This is a common strategy to express counter speech~\cite{DBLP:conf/icwsm/MathewSTRSMG019}. 

% Start: Table 
\begin{table*}[ht!]
	\small
	\centering
	
	\begin{tabular}{lcrcccp{0.2cm}}
		\toprule
		Authors                  & Source   & Size   &Labels                &Context?   &Counter? \\
		\midrule
		\citet{waseem-hovy-2016-hateful}    &Twitter   &1,607   &Sexism/Racism/Normal  &\xmark                   &\xmark       \\
		\citet{hateoffensive}     &Twitter   &24,783     &Hate/Offense/Neither  &\xmark                    &\xmark       \\
		\citet{nobata2016abusive}	     &Yahoo!	&2,000	 &Hate/Derogatory/Profanity/Clean  &\xmark          &\xmark      \\
		\citet{DBLP:conf/aaai/MathewSYBG021}  & Gab      &1,1093   &Hateful/Offensive/Normal &\xmark &\xmark \\
		\citet{gao-huang-2017-detecting}	     &Fox News	&1,528	 &Hateful/Non-hateful	&preceding comment      &\xmark \\
		\citet{qian-etal-2019-benchmark}	     &Reddit	&22,324	 &Hate/Non-hate	        &full conversation      &\xmark \\ 
		\citet{pavlopoulos-etal-2020-toxicity}  &Wikipedia	&20,000	 &Toxic/Non-toxic	    &preceding comment      &\xmark \\ %\addlinespace
		\citet{DBLP:journals/corr/abs-2103-14916}  &Twitter	&8,018	 &Abuse/Non-abuse	    &preceding comment      &\xmark \\ %\addlinespace
		\hdashline %\addlinespace
		\citet{DBLP:conf/icwsm/MathewSTRSMG019}	     &YouTube	&13,924	 &Non-counter/Counter (6,898)	&\xmark	        	  &\cmark \\
		\citet{he2021racism}  	     &Twitter	&2,400	 &Hate/Neutral/Counter (517) &\xmark		      &\cmark \\
		\citet{vidgen-etal-2021-introducing}	     &Reddit	&27,494	 & Taxonomy including Counter (220)	&full conversation	        	  &\cmark \\
		\textbf{Ours}            &Reddit    & 6,846  &Hate/Neutral/Counter (1,622) &  preceding comment     &\cmark \\    
		\bottomrule
	\end{tabular}
	\caption{Comparison of corpora with hate (above dashed line) and counter-hate annotations (below dashed line, some also include hate).
	  \citet{vidgen-etal-2021-introducing} is the only one considering counter-hate and context, but they only have 220 instances of counter hate.
	  Numbers between parenthesis indicate the number of counter-hate instances.
	  %Additionally, we investigate the role of context in the annotation process and ex
      %We study the role of context in the annotation and detection of hate and counter-hate in Reddit conversations and have 1,099 Counter-hate comments. 
    }
	%different public available datasets for hate speech and counter speech detection.}
	\label{t:datasets}
\end{table*}
% End: Table 

In this study, we answer the following questions:
\begin{compactenum}
	\item Does conversational context affect if a comment is perceived as Hate, Neutral, or Counter-hate by humans? (It does.) %the perceptions of a comment to be Hate or Counter-hate?
	\item Do models to identify Hate, Neutral, and Counter-hate benefit from incorporating context? (They do.)
	% Does adding context improve the performance of the classifiers?
\end{compactenum}

To answer the first question,
we create a collection of (\emph{Parent}, \emph{Target}) Reddit comments and annotate the \emph{Targets} with three labels (Hate, Neutral, Counter-hate)
in two independent phases:
showing annotators 
(a) only the \emph{Target}
or
(b) the \emph{Parent} and the \emph{Target}. 
We limit context to the parent comment.
While the full conversation could provide additional information,
it is also known to affect annotators' stance~\cite{dutta2020changing} and introduce biases.
%The setting of context is limited to one preceding comment, i.e., \emph{Parent}, partially because providing the full conversation may affect the annotators' stance~\cite{dutta2020changing} 
%and 
%distract them to judge from a relatively neutral perspective. 
%In addition, not all preceding comments are directly related to the \emph{Target}. 
%The number of preceding comments that is useful for the annotation and detection of the \emph{Target} differs case by case. 
%In this study, we use the limited conversational context to focus on verifying its effect. 
We find that human judgments are substantially different when the  \emph{Parent} is shown. 
Thus the task of annotating Hate and Counter-hate requires taking into account the context.

To answer the second question,
we experiment with context-unaware and context-aware classifiers to detect if a given \emph{Target} is Hate, Neutral, or Counter-hate. 
Results show that adding context does benefit the classifiers significantly.

In summary, the main contributions of this paper are:\footnote{Code and data available at \url{https://github.com/xinchenyu/counter\_context}}
\begin{inparaenum}[(a\upshape)]
	\item a corpus with 6,846 pairs of (\emph{Parent}, \emph{Target}) Reddit comments and annotations indicating whether the \emph{Target}s are Hate, Neutral, or Counter-hate;
	\item annotation analysis showing that the problem requires taking into account context, as the ground truth changes;
	\item corpus analysis detailing the kind of language people use to express Hate and Counter-hate;
	\item experiments showing that context-aware neural models obtain significantly better results;
	and
	\item qualitative analysis revealing when context is beneficial and the remaining errors made by the best context-aware model.
\end{inparaenum}

\section{Related Work}
Hate speech in user-generated content has been an active research area recently~\cite{fortuna2018survey}. 
Researchers have built several datasets for hate speech detection from diverse sources such as Twitter \cite{waseem-hovy-2016-hateful, hateoffensive}, Yahoo! \cite{nobata2016abusive}, Fox News \cite{gao-huang-2017-detecting}, Gab \cite{DBLP:conf/aaai/MathewSYBG021} and Reddit \cite{qian-etal-2019-benchmark}.

Compared to hate speech detection, few studies focus on detecting counter speech~\cite{DBLP:conf/icwsm/MathewSTRSMG019, garland-etal-2020-countering, he2021racism}. 
\citet{DBLP:conf/icwsm/MathewSTRSMG019} collect and hand-code 6,898 counter hate comments from YouTube videos targeting Jews, Blacks and LGBT communities. 
\citet{garland-etal-2020-countering} work with German tweets and define hate and counter speech based on the communities to which the authors belong.
\citet{he2021racism} use a collection of hate and counter hate keywords relevant to COVID-19 and create a dataset containing 359 counter hate tweets targeting Asians.
Another line of research focuses on curating datasets for counter speech generation using
crowdsourcing~\cite{qian-etal-2019-benchmark} or with the help of trained operators \cite{chung-etal-2019-conan, fanton-etal-2021-human}. 
However, synthetic language is rarely as rich as language in the wild.
Even if it were, conclusions and models from synthetic data may not transfer to the real world.
In this paper, we work with user-generated content expressing hate and counter-hate rather than synthetic content.
%We propose to retrieve Counter-hate samples from online discussions for further countering actions.

Table \ref{t:datasets} summarizes existing datasets for Hate and Counter-hate detection. 
Most of them do not include context information. In other words, the preceding comments are not provided when annotating \emph{Target}s. 
Context does affect human judgments and has been taken into account for Hate detection \cite{gao-huang-2017-detecting,pavlopoulos-etal-2020-toxicity,DBLP:journals/corr/abs-2103-14916,vidgen-etal-2021-introducing}.
\citet{gao-huang-2017-detecting} annotate hateful comments in the nested structures of %10
Fox News discussion threads. 
\citet{vidgen-etal-2021-introducing} introduce a dataset of Reddit comments with annotations in 6 categories taking into account context. 
However, the inter annotator agreement is low (Fleiss’ Kappa 0.267) and the number of Counter-hate instances is small (220).
Moreover, both studies use contextual information without identifying the role context plays in the annotation and detection.
\citet{pavlopoulos-etal-2020-toxicity} allow annotators to see one previous comment to annotate Wikipedia conversations.
%create their corpus by allowing one preceding comment to be seen when manually coding \textit{Targets} from Wikipedia conversations. 
They find context matters in the annotation but provide no empirical evidence showing whether models to detect toxicity benefit from incorporating context. 
%incorporating context have better performance in the detection of toxicity. 
\citet{DBLP:journals/corr/abs-2103-14916} re-annotate an existing corpus to investigate the role of context in abusive language. %annotation and detection.
They found context does matter. 
Utilizing conversational context has also been explored in text classification tasks such as 
sentiment analysis~\cite{ren2016context}, stance~\cite{ZUBIAGA2018273} and 
sarcasm~\cite{ghosh-etal-2020-report}. 
In this paper, we investigate the role of context in Hate and Counter-hate detection. 

\section{Dataset Collection and Annotation}
\label{s:datasets-Collection-Annotation}
We first describe our procedure to collect (\emph{Parent}, \emph{Target}) pairs, where both \emph{Parents} and \emph{Targets} are Reddit comments in English.
Then, we describe the annotation guidelines and the two annotation phases:
showing annotators (a) only the \emph{Target} and (b) the \emph{Parent} and \emph{Target}.
%We create two corpuses for Hate and Counter-hate detection: one with context, the other without.
The two independent phases allow us to quantify how often context affects the annotation of Hate and Counter-hate.

\subsection{Collecting (\emph{Parent}, \emph{Target}) pairs}
\label{ss:collecting-pairs}

%We focus on r/Mensrights, a subreddit in Reddit where men's perspectives, gripes, desires and frustrations are explicitly expressed.
%Women and feminism are the most common targets of hostility \cite{flood2004backlash,farrell2019exploring}. 
In this work, we focus on Reddit, a popular social media site. It is an ideal platform for data collection due to the large size of user populations and many diverse topics \cite{DBLP:conf/icwsm/BaumgartnerZKSB20}. We use a list of hate words to retrieve Reddit conversations to keep the annotation costs reasonable while creating a (relatively) large corpus of counter speech.  
We start with a set of 
1,726 hate words from two lexicons: 
Hatebase\footnote{\url{http://hatebase.org/}}
and 
a harassment corpus \cite{rezvan2018quality}. 
We remove ambiguous words % I thought that it is better to not give the example: we need space and there is really no good reason to discard some hateful comments targeting homosexuals.
%in different contexts
following \citet{DBLP:conf/icwsm/ElSheriefKNWB18}. 
%For example, ``fruits'' is included in Hatebase because it can be a derogatory term referring to homosexuals,
%although it generally refers to a food. 
To collect (\emph{Parent}, \emph{Target}) pairs, we use the following steps.
First, we retrieve comments containing at least one hate word~(\cwhw{}).
%that are potentially hateful and mark them as \textit{Current}. 
Second, we create a (\emph{Parent}, \emph{Target}) pair using \cwhw{} as \emph{Target} and its preceding comment as \emph{Parent}.
Third, we create a \textit{(Parent, Target)} pair using \cwhw{} as \emph{Parent} and each of its replies as \emph{Target}.
Lastly, we remove pairs if the same author posted the \emph{Parent} and \emph{Target}. 
%(ii) To examine if context affects annotations, for each \textit{Current}, we get the closest previous comment and denote them as \textit{Previous}. 
%(iii) As counter speech are usually generated against hateful comments, we fetch all direct replies to \textit{Current} and mark them as \textit{Next}. 
%The data are in the form of paired conversations as either \{\textit{Previous}, \textit{Current}\} or \{\textit{Current}, \textit{Next}\}. 
%(iv) We refer the first object in the bracket to \textit{Parent} and the second object as \textit{Target}. 
%Hence, both \{\textit{Previous}, \textit{Current}\} 
%and
% \{\textit{Current}, \textit{Next}\} are in the form of \{\textit{Parent}, \textit{Target}\}. 
%We set the maximum length of \emph{Target} to 20 tokens as 77\% of \emph{Target} labels affected by their \emph{Parents} were under 20 tokens in our pilot study. 
We retrieve 6,846 ({\emph{Parent}, \emph{Target}) pairs with PushShift~\cite{DBLP:conf/icwsm/BaumgartnerZKSB20} from 416 submissions. We also collect the title of the discussion from which each pair is retrieved.
	
\subsection{Annotation Guidelines}
To identify whether a \emph{Target} is Hate, Neutral, or Counter-hate, 
we crowdsource human judgments from non-experts.
Our guidelines reuse the definitions of Hate by \citet{ward1997free} and Counter-hate by \citet{DBLP:conf/icwsm/MathewSTRSMG019} and \citet{vidgen-etal-2021-introducing}:
	%We first provide a conversation in the form of a \{\textit{Parent}, \textit{Target}\} pair, 
	%and 
	%then ask the crowd-workers to pick one for \textit{Target} out of the three options:
	%based on intrinsic characteristics such as race, gender, body form, 
\begin{compactitem}
	\item \textbf{Hate}: the author attacks an individual or a group with the intention to vilify, humiliate, or incite hatred;
	\item \textbf{Counter-hate}: the author challenges, condemns the hate expressed in another comment, or calls out a comment for being hateful;
	\item \textbf{Neutral}: the author neither conveys hate nor opposes hate expressed in another comment.
\end{compactitem}

\noindent \textbf{Annotation Process}
We chose Amazon Mechanical Turk (MTurk) as the crowdsourcing platform. 
We replace user names with placeholders (User\_A and User\_B) owing to privacy concerns.
The annotations took place in two independent phases.
In the first phase, annotators are first shown the \emph{Parent} comment.
After a short delay, they click a button to show the \emph{Target} and then after another short delay they submit their annotation.
Delays are at most a few seconds and proportional to the length of the comments.
Our rationale behind the delays is to ``force'' annotators to read the \emph{Parent} and \emph{Target} in order.
In the second phase, annotators label each \emph{Target} without seeing the preceding \emph{Parent} comment.
A total of 375 annotators were involved in the first phase and 299 in the second phase.
There is no overlap between annotators thus we eliminated the possibility of biased annotators remembering the \emph{Parent} in the second phase.
	%When annotating each instance, the left panel will remain static, while the right panel will be dynamic: 
	%we first show crowd-workers the discussion title along with the comment by User\_A, 
	%after a few seconds delayed time a button will display 
	%and 
	%they need to click it to see the comment by User\_B. 
	%Following our rationale, crowd-workers are “enforced” to read \textit{Parent} first and then make the decision for \textit{Target}. 
	%The delayed time of the button is proportional to the number of tokens in the comment by User\_A. 

\noindent \textbf{Annotation Quality}
	Crowdsourcing may attract spammers~\cite{sabou-etal-2014-corpus}. 
	For quality control, we first set a few requirements for annotators:
		they must be located in the US
		and have a 95\% approval rate over at least 100 Human Intelligence Tasks (HITs). 
		We also block annotators 
		who submit more than 10 HITs with an average completion time below 5 seconds (half the time required in our pilot study).
		As the corpus contains vulgar words,
		we require annotators to pass the Adult Content Qualification Test. The reward per HIT is \$0.05.

	The second effort is to identify bad annotators and filter out their annotations until we obtain \textit{substantial} inter-annotator agreement.
	We collect five annotations per HIT and compute MACE~\cite[Multi-Annotator Competence Estimation]{hovy-etal-2013-learning} for each annotator. 
	MACE is devised to rank annotators by their competence and adjudicate disagreements based on annotator competence (not the majority label).
	%output best estimates by aggregating their annotations. 
	Then, we use Krippendorff’s  $\alpha$~\cite{krippendorff} to estimate inter-annotator agreement: $\alpha$ coefficients at or above 0.6 are considered \textit{substantial} (above 0.8 are considered \emph{nearly perfect})~\cite{artstein2008inter}. 
	We repeat the following steps until  $\alpha$  $\geq$ 0.6: 
\begin{compactenum}
		\item Use MACE to calculate the competence score of all annotators.
		\item Discard all the annotations by the annotator with the lowest MACE score.
		\item Check Krippendorff’s $\alpha$ on the remaining annotations. Go to (1) if $\alpha < 0.6$.
\end{compactenum}
	
	The final corpus consists of 6,846 (\emph{Parent}, \emph{Target}) pairs and a label assigned to each \emph{Target} (Hate, Counter-hate, or Neutral).
	The ground truth we experiment with (Section \ref{s:experiments}) is the label obtained taking into account the \emph{Parent} (first phase)}. The second phase, which disregards the \emph{Parent}, was conducted for analysis purposes~(Section~\ref{s:analysis}).
	We split the corpus into two subsets:
	(a) Gold (4,751 pairs with $\alpha \ge 0.6$) and
	(b) Silver (2,095 remaining pairs).
	As we shall see, the Silver pairs are useful to learn models.

\section{Corpus Analysis}
\label{s:analysis}

% Start: Table 
\begin{table}
	\small
	\centering
	
	\begin{tabular}{ccccc}
		\toprule
		\multicolumn{2}{c}{} & \multicolumn{3}{c}{Without \emph{Parent}}\\
		\cmidrule{3-5}
		&  & Hate & Counter-hate & Neutral\\
		\midrule
		\multirow{3}{*}{\begin{sideways}With\end{sideways}}
		&  Hate & 57.4 & 8.4 & 34.2\\
		&  Counter-hate & 18.7 & 26.2 & 55.1\\
		&  Neutral & 9.7 & 8.1 & 82.2\\
		\bottomrule
	\end{tabular}
	
	\caption{Confusion matrix (percentages) showing annotation changes depending on whether annotators are shown the \emph{Parent} of the \emph{Target} comment.}
	%	to show label percentage 
	%		(e.g., 57.4\% \textit{Target} are labeled as Hate whether provide \textit{Parent} or not. 
	%		34.2\% Hate changed to Neutral without \textit{Parent}).
	\label{t:confusion-matrix}
\end{table}
% End: Table 

% Start: Table 
\begin{table}[t!]
	\small
	\centering
	
	\begin{tabular}{p{4.5cm}  c c}
		\toprule
		Example & With & Without \\
		\midrule
		\emph{Parent}: That chick needs a high-five in the face with a chair. Damn her for making us look bad! &  & \\
		\emph{Target}: A brick is more effective.  &Hate &  Neutral\\
		\midrule
		\emph{Parent}: If I knew her I would sh*t in her mailbox. &  & \\
		\emph{Target}: The poor mail carrier in that neighborhood doesn't deserve that.  &Counter &  Neutral\\
		\midrule
		\emph{Parent}: Go watch your incest porn on your own time. &  & \\
		\emph{Target}: You're a sick person. & Counter &  Hate\\
		\bottomrule
	\end{tabular}
	
	\caption{Examples of \emph{Target} comments whose labels change depending on whether annotators are shown the \emph{Parent}
	  of the \emph{Target} comment (with and without).}
	\label{t:corpus-examples}
\end{table}
% End: Table 

% Start: Table 
\begin{table*}
	\centering
	\small
	\begin{tabular}{lcccccccc}
		\toprule
		\multicolumn{1}{c}{} & \multicolumn{3}{c}{\textit{Title}} & \multicolumn{3}{c}{\emph{Parent}} & \multicolumn{2}{c}{\emph{Target}}\\
		\cline{2-3}  \cline{5-6}  \cline{8-9}
		\multicolumn{1}{c}{} & p-value & Bonferroni & & p-value & Bonferroni & & p-value & Bonferroni\\
		\hline
		Textual factors \\
		~~~~Total tokens    &$\downarrow\downarrow$ &\xmark & &$\uparrow\uparrow\uparrow$ & \cmark & & & \\
		~~~~Question marks  &   &  &     &   &  &  & $\uparrow\uparrow\uparrow$ &\cmark \\
		~~~~1st person pronouns     &   &   &    &$\downarrow\downarrow\downarrow$ & \cmark & & & \\
		~~~~2nd person pronouns     &   &   &    &$\uparrow\uparrow\uparrow$   &\cmark & &$\uparrow\uparrow$ & \xmark\\
		\hline
		Sentiment and cognitive factors \\
		~~~~Profanity words & & & &$\uparrow\uparrow\uparrow$ &\cmark & & $\downarrow\downarrow\downarrow$ & \cmark \\
		~~~~Problem-solving words & & & & & & &$\uparrow\uparrow\uparrow$   &\cmark \\
		~~~~Awareness words & & & & & & &$\uparrow\uparrow\uparrow$   &\cmark \\
		~~~~Negative words  & $\downarrow$ &\xmark & &$\uparrow\uparrow\uparrow$ &\cmark & &$\downarrow\downarrow\downarrow$ & \cmark \\
		~~~~Disgust words & & & & & & & $\downarrow\downarrow\downarrow$  & \cmark \\
		~~~~Enlightenment words  & & & & & & & $\uparrow\uparrow\uparrow$ & \cmark \\
		~~~~Conflicting words &$\downarrow\downarrow\downarrow$ &\cmark & & & & & &\\ 
		\bottomrule
	\end{tabular}
	\caption{Linguistic analysis comparing the \textit{Titles}, \emph{Parents} and \emph{Targets} in Counter-hate and Hate \emph{Target} comments. 
		Number of arrows indicate the p-value (t-test; one: p\textless 0.05, two: p\textless 0.01, and three: p\textless 0.001).
		Arrow direction indicates whether higher values correlate with Counter-hate (up) or Hate (down). 
		A check mark (\cmark) indicates that the test passes the Bonferroni correction.}
	\label{t:linguistic-analysis}
\end{table*}
% End: Table 

\paragraph{Does conversational context affect if a comment is perceived as Hate or Counter-hate?}
Yes, it does. 
Table \ref{t:confusion-matrix} presents the percentage of labels that change and remain the same depending on whether annotators are shown the \emph{Parent}, i.e., the context.
%remains same and changes to others labels when \textit{Parent} is not provided.
Many \emph{Target} comments that are perceived as Hate or Counter-hate become Neutral (34.2\% and 55.1\% respectively) when the \emph{Parent} is provided. 
More surprisingly, many \emph{Target} comments are perceived with the opposite label (from Hate to Counter-hate (8.4\%) or from Counter-hate to Hate (18.7\%)) when the \emph{Parent} comments are shown.
%A lot of disagreements are made when crowd-workers mistake Hate or Counter-hate as Neutral when \textit{Parent} is not provided (34.2\% and 55.1\% respectively).

We show examples of label changes in Table \ref{t:corpus-examples}. 
In the first example, annotators identify the \emph{Target} (``A brick is more effective.'') as Neutral without seeing the \emph{Parent}.
In fact, a female is the target of hate in the \emph{Parent}, and the author of \emph{Target} replies with even more hatred (and the ground truth label is Hate).
In the second example, the \emph{Target} alone is insufficient to tell if it is Counter-hate. 
When annotators see the \emph{Parent}, however, they understand the author of \emph{Target} counters the hateful content in the \emph{Parent} by showing empathy towards the mail carrier. 
In the last example, the \emph{Target} alone is considered Hate because it attacks someone by using the phrase ``sick person''.
When the \emph{Parent} is shown, however, the annotators understand the \emph{Target} as calling out the \emph{Parent} to be inappropriate. 
 
\begin{figure}[t]
	\centering
	\includegraphics[width=0.75\linewidth]{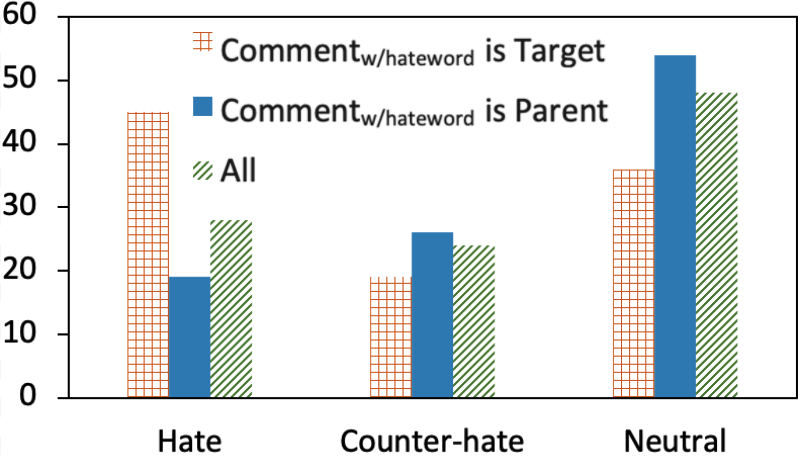}
	\caption{Label distribution in \emph{Target}s depending on whether \cwhw{} is the \emph{Parent} or the \emph{Target}.}
	\label{fig:label_distribution}
\end{figure} 

% Start: Table 
\begin{table*}[ht!]
	\small
	\centering
	
	\begin{tabular}{l ccc ccc ccc ccc}
		\toprule
		\multicolumn{1}{c}{} & \multicolumn{3}{c}{Hate} & \multicolumn{3}{c}{Counter-hate} & \multicolumn{3}{c}{Neutral} & \multicolumn{3}{c}{Weighted Average} \\
		\cmidrule(lr){2-4} \cmidrule(lr){5-7} \cmidrule(lr){8-10} \cmidrule(lr){11-13} 
		& P & R & F1 & P & R & F1 & P & R & F1 & P & R & F1 \\ 
		\hline \addlinespace
		Majority Baseline & 0.00 & 0.00 & 0.00 & 0.00 & 0.00 & 0.00 & 0.51 & 1.00 & 0.67 & 0.26 & 0.51 & 0.34 \\ \addlinespace
		%		& & & & & & & & & & & & & & &\\
		
		Trained with Target & 0.56 & 0.55 & 0.56 & 0.41 & 0.36 & 0.38 & 0.67 & 0.71 & 0.69 & 0.58 & 0.59 & 0.58\\
		~~~~~~~~+ Silver & 0.58 & 0.55	&0.57&0.44	&0.42	&0.43&0.69	&0.72	&0.70&0.60	&0.61	&0.61\\
		~~~~~~~~+ Related task & 0.56 & 0.55	&0.56&0.51	&0.41	&0.45&0.68	&0.74	&0.71&0.61	&0.61	&0.61\\
		~~~~~~~~+ Silver + Related task & 0.55 & 0.56 & 0.56 & 0.49 & 0.53 & 0.51 & 0.67 & 0.69 & 0.70 & 0.61 & 0.61 & 0.61\\ \addlinespace
		Trained with Parent\_Target & 0.56 & 0.62 & 0.59 & 0.52 & 0.38 & 0.44 & 0.68 & 0.72 & 0.70 & 0.61 & 0.62 & 0.61\\
		~~~~~~~~+ Silver† & 0.58	&0.57	&0.57	&0.49	&0.51&	0.50	&\textbf{0.72}	&\textbf{0.71}	&\textbf{0.72}	&0.63	 &0.63	&0.63 \\
		~~~~~~~~+ Related task† & \textbf{0.55} & \textbf{0.66} & \textbf{0.60} & 0.54 & 0.43 & 0.48 & 0.71 &0.70 & 0.71 & 0.63 & 0.63 & 0.63  \\
		~~~~~~~~+ Silver + Related task‡ & 0.55 & 0.65 & 0.60 & \textbf{0.54} & \textbf{0.52} & \textbf{0.53}& 0.74 &0.68 & 0.71 & \textbf{0.64} & \textbf{0.64} & \textbf{0.64} \\
		\bottomrule
		
	\end{tabular}
	\caption{Results obtained with several systems. 
		We indicate statistical significance (McNemar’s test \cite{mcnemar1947note} over weighted average)
		with respect to the model trained with the \emph{Target} only using neither Silver nor pretraining on related tasks as follows: 
		† indicates $p<0.05$ and ‡ indicates $p<0.01$.
		Training with the \emph{Parent} and \emph{Target}
		coupled with blending Silver annotations and pretraining with stance corpora yields the best results.
		The supplementary materials detail the results pretraining with all related tasks we consider.}
	\label{t:model-results}
\end{table*}
% End: Table 

\paragraph{Label distribution and linguistic insights} Figure~\ref{fig:label_distribution}  shows the label distribution for all pairs (rightmost column in each block)
and for pairs in which \cwhw{} (i.e., the comment containing at least one hate word) is the \emph{Parent} or \emph{Target}.
The most frequent label assigned to \emph{Target} comments is Neutral~(49\%) followed by Hate~(28\%) and Counter-hate~(23\%).
While \emph{Target} comments containing a hate word are likely to be Hate (45\%), some are Counter-hate (19\%) with context.

%: most \textit{Target}s are Neutral (48\%), followed by 28\% as Hate and 24\% as Counter-hate. 
%We also show the percentage of each label where \textit{Current} is either as \textit{Target} or \textit{Parent} in \{\textit{Parent}, \textit{Target}\}. 
%We observe a few interesting findings: 
%comments containing hate words are most likely to be hateful (45\% of \textit{Current}s are Hate); 
%counter hate speech might also include hate words (19\% of \textit{Current}s are Counter-hate).

We analyze the linguistic characteristics of \textit{Title}s, \emph{Parent}s and \emph{Target}s when the \emph{Target}s are Hate or Counter-hate with context to shed light on the differences between the language people use in hate and counter speech. 
We combine the set of hate words with profanity words to check for profanity words.\footnote{\url{https://github.com/RobertJGabriel/google-profanity-words-node-module/blob/master/lib/profanity.js}} 
We analyze sentiment and cognitive factors using the Sentiment Analysis and Cognition Engine (SEANCE) lexicon,
a popular tool for psychological linguistic analysis \cite{crossley2017sentiment}. 
Statistical tests are conducted using unpaired t-tests between the groups, of which the \emph{Target}s are Counter-hate or Hate (Table \ref{t:linguistic-analysis}). 
We also report whether each feature passes the Bonferroni correction as multiple hypothesis tests are performed. 
We draw several interesting insights:
\begin{compactitem}
	\item Questions marks in \emph{Target} signal Counter-hate.
	They are often rhetorical questions.
	%We observe that people are inclined to use rhetorical questions as a way to counter hateful comments. 
	\item Fewer 1st person pronouns (e.g., I, me) and more 2nd person pronouns (e.g., you, your) in the \emph{Parent} signal that the \emph{Target} is more likely to be Counter-hate. This is due to the fact that people tend to target others in hateful content. 
	\item High profanity count in the \emph{Parent} signals that the \emph{Target} is Counter-hate, while high profanity count in the \emph{Target} signals Hate. 
	\item More words related to awareness, enlightenment and problem-solving in the \emph{Target} signal Counter-hate. 
	\item When there are more negative words in the \emph{Parent}, the \emph{Target} tends to be Counter-hate.
	\emph{Target}s labeled as Counter-hate contain fewer negative and disgusting words.
\end{compactitem}

\section{Experiments and Results} \label{s:experiments}
We build neural network models to identify if a \emph{Target} comment is Hate, Counter-hate, or Neutral. 
We randomly split Gold instances (4,751) as follows: 70\% for training, 15\% for validation and 15\% for testing. 
Silver instances are only used for training.

\paragraph{Neural Network Architecture} We experiment with neural classifiers built on top of the RoBERTa transformer \cite{DBLP:journals/corr/abs-1907-11692}. 
The neural architecture consists of a pretrained RoBERTa transformer, 
a fully connected layer (768 neurons and Tanh activation), 
and 
another fully connected layer (3 neurons and softmax activation) to make predictions (Hate, Counter-hate, or Neutral). 
To investigate the role of context, we consider two textual inputs:
\begin{compactitem}
	\item the \textit{Target} alone (Target), and 
	\item the \textit{Parent} and the \textit{Target} (Parent\_Target).
\end{compactitem}
We concatenate the \textit{Target} and the \textit{Parent} with the [SEP] special token. % when feed them both to the network. 
We conduct multiple runs of experiments, which show consistent results. The hyperparameters and other implementation details are presented in the Appendix. 
We also experiment models that take the title of the discussion as part of the context, 
but it is not beneficial.
%We have tried to include the titles of the Reddit discussions but discard them as they do not yield better results.

We implement two strategies to enhance the performance of neural models:

\paragraph{Blending Gold and Silver}
We adopt the method by \citet{shnarch-etal-2018-will} to determine whether Silver annotations are beneficial.
%identify if the use of Silver data may assist model training. 
There are two phases in the training process: 
\textit{m} blending epochs using all Gold and a fraction of Silver, and then \textit{n} epochs using all Gold. 
In each blending epoch, Silver instances are fed in a random order to the network. 
The fraction of Silver is determined by a blending factor \(\alpha\) \(\in\) [0..1]. 
The first blending epoch is trained with all Gold and all Silver, 
and 
the amount of Silver to blend is reduced by $\alpha$ in each epoch.
%Eventually, there is no Silver in the training data and we use all Gold to train the remaining \textit{n} epochs.  

\paragraph{Pretraining with Related Tasks} 
We also experiment with several corpora to investigate whether pretraining with related tasks is beneficial. 
Specifically, we pretrain our models with existing corpora annotating: 
(1) hateful comments: hateful or not hateful \cite{qian-etal-2019-benchmark}, 
and 
hate speech, offensive, or neither \cite{hateoffensive}; 
(2) sentiment: negative, neutral, or positive \cite{rosenthal-etal-2017-semeval}; 
(3) sarcasm: sarcasm or not sarcasm \cite{ghosh-etal-2020-report}; 
and 
(4) stance: agree, neutral, or attack \cite{pougubiyong2021debagreement}.

% Start: Table 
\begin{table*}
	\small
	\centering
	
	\begin{tabular}{@{\hspace{.03in}}p{2.5cm}p{0.1cm}p{0.6cm}p{7.6cm}ll@{\hspace{.03in}}}
		\toprule
		Error Type & \% & \multicolumn{2}{l}{Example} & Parent\_Target  & Target  \\ \midrule
		
		Lack of information & 48 & \emph{Parent}: & Women can hover..? &  & \\
		&    & \emph{Target}: & No, they can't, but for some reason they keep trying and it gets sh*t everywhere. & Hate & Neutral  \\
		\midrule
		Negation   & 27 & \emph{Parent}: & It's a joke you pu**y. &  &  \\
		&    & \emph{Target}: & I don't see sexism as a joke, especially on a site dedicated to calling out sexism. & Counter-hate & Neutral \\ 
		\midrule
		Sarcasm or irony & 19 & \emph{Parent}: & You must have been a real baller banging out those eighth graders as a High School senior. &  &  \\
		&    & \emph{Target}: & Glad to see you have no rational argument left except childish jokes.  We're done here pal.  & Counter-hate & Hate \\
		\midrule
		\multirow{2}{1in}{Hate without swear words} & 8 & \emph{Parent}: & Name a dildo `misogyny' so you can *literally* internalize it. &  & \\
		&    & \emph{Target}: & lol. Misogyny can already turn me on so that's a good idea. & Hate & Neutral  \\
		\bottomrule
	\end{tabular}
	\caption{Most common error types made by the \textit{Target} only network (Target) that are fixed by the context-aware neural network (Parent\_Target).
		% coms fixed by PT-Target compared to Target.
	}
	\label{t:error-target}
\end{table*}
% End: Table 

\subsection{Quantitative Results}
We present results with the test split in Table \ref{t:model-results}. 
The majority baseline always predicts Neutral.
%The top row presents the results with the majority baseline (i.e., always Neutral).
The remaining rows present the results with the different training settings:
training with the \emph{Target} or both the \emph{Parent} and \emph{Target};
training with only Gold or blending Silver annotations;
and
pretraining with related tasks.
We provide here results pretraining with the most beneficial task, stance detection,
and present additional results in the Appendices.
% of training depending on whether blending Silver instances or pretraining with related tasks are used when the input is Target and Parent\_Target respectively.
Blending Gold and Silver annotations requires tuning $\alpha$.
We did so empirically using the training and validations splits, like other hyperparameters.
We found the optimal value to be 0.3 when blending Silver (+ Silver rows) and 1.0 when blending Silver and pretraining with a related task (+Silver + Related task rows). 
%In the blending of Gold and Silver data, the value of \(\alpha\) is empirically determined. 
%We choose \(\alpha\) as 0.3 for Target+Silver and Parent\_Target+Silver, 
%and 
%\(\alpha\) as 1 when using both Silver and pretraining with related tasks as it generates the best results. 

As shown in Table \ref{t:model-results}, blending Gold and Silver annotations obtains better results (F1 weighted average) than using only Gold (Target: 0.61 vs. 0.58; Parent\_Target: 0.63 vs. 0.61).
We also find that models pretrained for stance detection obtain better results than pretrained with other tasks (see detailed results in the Appendices).
Pretraining with stance detection data benefits models trained without context (Target: 0.61 vs. 0.58) and models with context (Parent\_Target: 0.63 vs. 0.61). 
These results indicate that stance information between \emph{Parent} and \emph{Target} is useful to determine whether the \emph{Target} is Hate, Counter-Hate or Neutral.
%These results indicate that these models successfully transfer knowledge about stance between \emph{Parent} and \emph{Target} into the task of detecting whether the \emph{Target} is Hate, Counter-Hate or Neutral.
%of the relationship between two comments in stance detection to our own tasks. 

We make two observations about the results obtained using neither strategy. %of the two strategies, we observe:
First, using the \emph{Target} alone obtains much better results than the majority baseline (0.58 vs. 0.34).
In other words, modeling the \emph{Target} allows the network to identify \emph{some} instances of Hate and Counter-hate despite the ground truth requires the \emph{Parent}.
Second, incorporating the \emph{Parent} comment is beneficial: the F1 score for all classes are higher (Hate: 0.59 vs. 0.56, Counter-hate 0.44 vs. 0.38, Neutral 0.70 vs. 0.69), and so is the weighted average (0.61 vs. 0.58). 
The findings are consistent (weighted F1) using either strategy (+Silver: 0.63 vs 0.61, +Related task: 0.63 vs 0.61) or both (0.64 vs. 0.61). 
The F1 scores of the three classes with Parent\_Target models are equal to or better than those by Target only models. 

Finally, the network (a) blending Gold and Silver annotations and (b) pretraining with stance detection achieves the best performance (Parent\_Target+Silver+Related task: 0.64). 
This result is statistically significant ($p<0.01$) compared to Target only model without blending Silver or pretraining with related tasks.

\begin{table*}
	\small
	\centering
	
	\begin{tabular}{p{2.5cm}p{0.1cm}p{0.6cm}p{6.7cm}ll}
		\toprule
		Error Type & \% & \multicolumn{2}{l}{Example} & Ground Truth & Predicted \\
		\midrule
		Negation & 28 & \emph{Parent}: & Those damn f**king white males, ruining it for everyone else. I'm going to a corner to process my guilt. &  & \\
		&    & \emph{Target}: & Don't forget male isn't a gender, it's a disease. & Hate & Counter-hate  \\
		\midrule
		Rhetorical question  & 27 & \emph{Parent}: & Men are the ones that made inequality.  &  &  \\
		&    & \emph{Target}: & Do you get paid to be a dumba** in the internet? & Hate & Counter-hate\\  
		\midrule
		Hate without swear & 8 & Parent: & Circumcision is good for men. &  & \\
		words &    & \emph{Target}: & Cut off the clitoris of women and cut of their breasts because of breast cancer then. & Hate & Neutral\\
		\midrule
		Non-hate with swear words & 8 & \emph{Parent}: & \textless I wonder if feminists ever consider that? No. They are b**ches incapable of empathy. &  &  \\
		&    & \emph{Target}: & This is the sh*t that gets screen capped and spread around to give this sub a bad name.  & Counter-hate & Hate  \\
		\midrule
		Intricate text   & 7 & \emph{Parent}: &Ah it's this again, f**king her and her cronies.  &  &  \\
		&    & \emph{Target}: & I have lost all respect for her. & Neutral & Hate \\
		\bottomrule
	\end{tabular}
	\caption{Most common errors made by the best context-aware network (predictions by Parent\_Target+Silver+Related task) compared to the ground truth.}
	\label{t:error-context}
\end{table*}
% End: Table 

\section{Qualitative Analysis}
When is adding the context beneficial? 
When does our best model make mistakes? 
To investigate these questions, we %perform a qualitative analysis. % of the errors made by each model.
%In particular, we
manually analyze the following:
\begin{compactitem}
	\item The errors made by the \emph{Target} only network that are fixed by the context-aware network (Trained with Parent\_Target, Table \ref{t:error-target}).
	\item The errors made by the context-aware network pretrained on related task (stance) and blending Silver annotations (Parent\_Target+Silver+Related task, Table~\ref{t:error-context}).	
\end{compactitem}

\paragraph{When does the context complement \emph{Target}?}
We analyze the errors made by the network using only the \emph{Target}
that are fixed by the context-aware network (Trained with Parent\_Target). 
Table \ref{t:error-target} exemplifies the most common error types.

The most frequent type of error fixed by the context-aware model is when there is \emph{Lack of information} in the \emph{Target} (48\%).
In this case, the \emph{Parent} comment is crucial to determine the label of the \emph{Target}.
In the example, knowing what the author of \emph{Target} refers to (i.e., a rhetorical question, \emph{Women can hover?}) is crucial to determine that the \emph{Target} is humiliating women as a group.
\begin{comment}
the \textit{Target} alone is not sufficient, 
and 
the concatenation of both the \textit{Target} and the \textit{Parent} provides key information. 
Considering the first conversation in Table \ref{t:error-target} as an example, the \textit{Target} alone is treated as a neutral reply until provided with the \textit{Context}, which points out women as the target of attack. 
Without the \textit{Parent}, the network Target fails to make the connection, but that adding the Parent\_Target complements it and predicts that the reply is a hateful comment humiliating women as a group.
\end{comment}

The second most frequent error type is \emph{Negation} (27\%).
In the example in Table \ref{t:error-target}, taking into account the \emph{Parent} allows the context-aware network to identify that the author of the \emph{Target} is scolding the author of \emph{Parent} and thus countering hate.
%The second most common error type is is due to the use of negation (27\%). 
%Taking account into the \textit{Context} allows the network to 
%address the object of negation ("joke" in this case) 
%and 
%identify the \textit{Target} as a Counter-hate comment against the hateful \textit{Parent}. 

\citet{nobata2016abusive} and \citet{qian-etal-2019-benchmark} have pointed out that sarcasm and irony make detecting abusive and hateful content difficult. 
We find evidence supporting this claim.
We also discover that by incorporating the \emph{Parent} comment, a substantial amount of these errors are fixed.
Indeed, 19\% of errors fixed by the context-aware network include sarcasm or irony in the \emph{Target} comment.

%In the third example, taking account into Parent\_Target allows the network to treat the \textit{Target} as Counter-hate using sarcasm as a way to rebut the user in the \textit{Parent}. 

Finally, the context-aware network taking into account the \emph{Parent}
fixes many errors (8\%) in which the \emph{Target} comment is Hate despite it does not contain swear words.
In the example, the \emph{Target} is introducing additional hateful content, which can be identified by the context-aware model when the \emph{Parent} information is used. 

%context-aware network can also help to predict the \textit{Target} as Hate even it has no swear words: the \textit{Target} contains ingrained prejudice against women  (8\%) (``Misoginy can already turn me on [...]"). 

% Start: Table 

\paragraph{When does the best model make errors?}
In order to find out the most common error types made by the best model (context-aware, Parent\_Target+Silver+Related task),
we manually analyze 200 random samples in which the output of the network differs from the ground truth.
%random conversations out of all the mistakes made by our best model (Training w/ Gold + Silver: Target + Parent\_Target). 
Table \ref{t:error-context} shows the results of the analysis.

Despite 27\% of errors fixed by the context-aware network (i.e., taking into account the \emph{Parent}) include negation in the \emph{Target},
\emph{negation} is the most common type of errors made by our best network~(28\%).
The example in Table \ref{t:error-context} is especially challenging as it includes a double negation.
%Negation in the \emph{Target} is the most common error type mada common error
%Although adding the \textit{Context} can make up for a few errors, understanding negation is still a tough task (28\%). 
%As shown in the corresponding example, the network fails to correctly predict the \textit{Target} as Hate when 
%the \textit{Target} uses double negations, 
%and 
%they negate different objects. 

We observe that \emph{Rhetorical questions} are almost as common (27\%).
This finding is consistent with the findings by \citet{schmidt-wiegand-2017-survey}.  
In the example, the best model fails to realize that the \emph{Target} is hateful, as it disdains the author of \emph{Parent}.
%Using rhetorical or suggestion questions in the \textit{Target} has comprised of 27\% errors in our sample, which in accordance with 
%Even when taking into account the \textit{Context}, the full network fail to refer "trash" to the user posting the \textit{Parent}. 

Swear words are present in a substantial number of errors.
Wrongly predicting a \emph{Target} without swear words as Counter-hate or Neutral accounts for 8\% of errors,
and
wrongly predicting a \emph{Target} with swear words as Hate accounts for another 8\% of errors.
%We also found that \emph{Target} comments
%(a)~sear words
%Falsely predicting 
%the \textit{Target} not containing swear words as Counter-hate/Neutral (8\%) 
%or 
%the \textit{Target} containing swear words as Hate (8\%) 
%is also a reason for the full network to make mistakes. 
As pointed out by \citet{hateoffensive}, hate speech may not contain hate or swear words. 
And vice versa, comments containing swear words may not be hateful \cite{zhang2019hate}.

Finally, we observe \emph{Intricate text} in 7\% errors. % are made because of the intricate text. 
Our best model identifies the \emph{Target} (``I have lost all respect for her.'') as Hate probably because by identifying the agreeing stance on the \emph{Parent}. Indeed, the author of \emph{Target} expresses his/her attitude without vilifying others. 
Hence, the ground truth label is Neutral. 

\section{Conclusions and Future Work}
Conversational context does matter in Hate and Counter-hate detection.
We have demonstrated so by 
(a) analyzing whether humans perceive user-generated content as Hate or Counter-hate
depending on whether we show them the \emph{Parent} comment
and
(b) investigating whether neural networks benefit from incorporating the \emph{Parent}.
We find that 38.3\% of human judgments change when we show the \emph{Parent} to annotators.
Experimental results demonstrate that networks incorporating the \emph{Parent} yield better results. 
Additionally, we show that noisy instances (Silver data) and pretraining with relevant datasets improves model performance. 
We have created and released a corpus of 6,846 (\emph{Parent}, \emph{Target}) pairs of Reddit comments with the \emph{Target} annotated as Hate, Neutral or Counter-hate.

Our work have several limitations. 
First, we only consider context as the parent comment.
While considering additional context might be sometimes beneficial,
doing so would require careful design to not bias annotations~\cite{dutta2020changing}. 
%First, we consider one preceding comment as the conversational context, while more preceding comments might be beneficial for the detection of some posts. 
Our research agenda includes exploring reliably strategies to consider more context and identify which parts are most important.
%Future studies will be conducted to include more contextual information and identify which part of the context matters in the annotation and detection. 
Second, people may have different opinions about what constitutes hate and counter speech due to different tolerances in online aggression.
%on whether a post is hate speech or not due to different tolerance in online aggression. 
We obtained the ground truth according to annotators' reliability (MACE scores),
which may lead to controversial samples falling in the Silver set and thus being considered only for training (not for testing).
%Here we assign a label according to the majority annotators' opinions, which may lead controversial samples to fall in the Silver set and not been evaluated. 
Finally, the keywords sampling used to create our corpus may introduce biases.
Despite we partially mitigate the issue by considering hateful comments in both the \emph{Parent} and \emph{Target}, community-based sampling~\cite{vidgen-etal-2021-introducing} could be applied in our future work.

\section{Ethical Considerations}
We use the PushShift API to collect data from Reddit.\footnote{\url{https://pushshift.io/api-parameters/}} Our collection process is consistent with Reddit's Terms of Service. The data are accessed through the data dumps on Google's BigQuery using Python.\footnote{\url{https://pushshift.io/
	using-bigquery-with-reddit-data/}} 

Reddit can be considered a public space for discussion which differs from a private messaging service \cite{vidgen-etal-2021-introducing}. 
Users consent to have their data made available to third parties including academics when they sign up to Reddit. 
Existing ethical guidelines state that in this situation explicit consent is not required from each user \cite{DBLP:conf/tto/ProcterWBHEWJ19}. 
We obfuscate user names as User\_A or User\_B to reduce the possibility of identifying users. In compliance with Reddit's policy, we would like to make sure that our dataset will be reused for non-commercial research only.\footnote{\url{https://www.reddit.com/wiki/api-terms/}}

The Reddit comments in this dataset were annotated by annotators using Amazon Mechanical Turk. 
We have followed all requirements introduced by the platform for tasks containing adult content. 
A warning was added in the task title. Annotators need to pass the Adult Content Qualification Test before working on our tasks. 
Annotators were compensated on average with \$8 per hour. We paid them regardless of whether we accepted their work.
Annotators' IDs are not included in the dataset.

\section*{Acknowledgements}
This work was supported by Research Seed Grant of the UNT College of Information.
We would like to thank the anonymous reviewers for their insightful comments and suggestions.

\bibliography{anthology,custom}
\bibliographystyle{acl_natbib}

\appendix

\section{Annotation Interface}
\label{sec:appendix}
\begin{figure*}[ht!]
	\centering
	\includegraphics[width=0.8\linewidth]{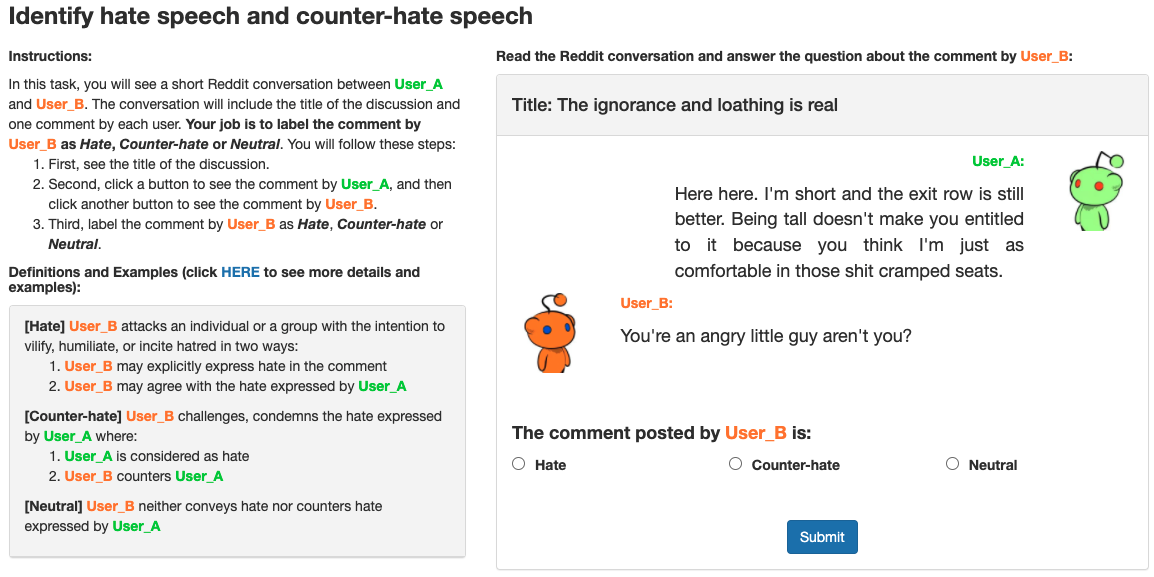}
	\caption{Screenshot of the annotation interface.
		The left panel displays the instructions and examples.
		The right panel displays the \emph{Parent} and the \emph{Target} to be annotated.}
	\label{fig:interface_context}
\end{figure*}

We show a screenshot of the annotation interface in Figure~\ref{fig:interface_context}.
% Start: Table 
\begin{table*}[ht!]
	\small
	\centering
	
	\begin{tabular}{l ccc ccc ccc ccc}
		\toprule
		\multicolumn{1}{c}{} & \multicolumn{3}{c}{Hate} & \multicolumn{3}{c}{Counter-hate} & \multicolumn{3}{c}{Neutral} & \multicolumn{3}{c}{Weighted Average} \\
		\cmidrule(lr){2-4} \cmidrule(lr){5-7} \cmidrule(lr){8-10} \cmidrule(lr){11-13} 
		& P & R & F1 & P & R & F1 & P & R & F1 & P & R & F1 \\
		\hline \addlinespace 
		Majority Baseline & 0.00 & 0.00 & 0.00 & 0.00 & 0.00 & 0.00 & 0.51 & 1.00 & 0.67 & 0.26 & 0.51 & 0.34 \\
		%		& & & & & & & & & & & & & & &\\
		
		Trained with ...\\
		~~~~Target & 0.56 & 0.55 & 0.56 & 0.41 & 0.36 & 0.38 & 0.67 & 0.71 & 0.69 & 0.58 & 0.59 & 0.58\\
		~~~~~~~~+ Hate\_Twitter &0.58	&0.53	&0.55	&0.46	&0.07	&0.12	&0.61	&0.88	&0.72	&0.57	&0.6	&0.54\\
		~~~~~~~~+ Hate\_Reddit & 0.57 & 0.52	&0.55&0.44	&0.32	&0.37&0.64	&0.75	&0.69&0.58	&0.59 &0.58\\
		~~~~~~~~+ Sentiment & 0.59 & 0.47	&0.53&0.00	&0.00	&0.00&0.59	&0.92	&0.72&0.45	&0.59	&0.50\\
		~~~~~~~~+ Sarcasm & 0.59 & 0.51 & 0.55 & 0.50 & 0.04 & 0.08 & 0.59 & 0.51 & 0.55 & 0.57 & 0.58 & 0.51\\
		~~~~~~~~+ Stance & 0.56 & 0.55 & 0.56 & 0.51 & 0.41 & 0.45 & 0.68 & 0.74 & 0.71 & 0.61 & 0.61 & 0.61\\
		Trained with ...\\
		~~~~Parent\_Target & 0.55 & 0.62 & 0.59 & 0.52 & 0.38 & 0.44 & 0.68 & 0.72 & 0.70 & 0.61 & 0.62 & 0.61\\
		~~~~~~~~+ Hate\_Twitter&0.49	&0.64	&0.56	&0.29	&0.13	&0.18	&0.66	&0.73	&0.7	&0.53	&0.57	&0.54\\
		~~~~~~~~+ Hate\_Reddit & 0.55	&0.64	&0.59	&0.48	&0.33&	0.39	&0.69	&0.73	&0.71	&0.61	 &0.62	&0.61 \\
		~~~~~~~~+ Sentiment & 0.53 & 0.59 & 0.56 & 0.40 & 0.23 & 0.29 & 0.68 &0.77 & 0.72 & 0.57 & 0.60 & 0.58  \\
		~~~~~~~~+ Sarcasm & 0.56 & 0.54 & 0.55 & 0.45 & 0.09 & 0.15& 0.62 &0.86 & 0.72 & 0.56 & 0.60 & 0.54 \\
		~~~~~~~~+ Stance & 0.55 & 0.66 & 0.60 & 0.54 & 0.43 & 0.48& 0.71 &0.70 & 0.71 & 0.63 & 0.63 & 0.63 \\
		\bottomrule
		
	\end{tabular}
	\caption{Detailed results (P, R, and F) predicting whether the \emph{Target} is Hate, Neutral or Counter-hate when the input is only the Target or the Parent\_Target. These results are using RoBERTa and pretrained with each related task. This table complements Table \ref{t:model-results} in the paper.}
	\label{t:detailed-results}
\end{table*}
% End: Table 

% Start: Table 
\begin{table*}[ht!]
	\small
	\centering
	
	\begin{tabular}{l ccc ccc ccc ccc}
		\toprule
		\multicolumn{1}{c}{} & \multicolumn{3}{c}{Hate} & \multicolumn{3}{c}{Counter-hate} & \multicolumn{3}{c}{Neutral} & \multicolumn{3}{c}{Weighted Average} \\
		\cmidrule(lr){2-4} \cmidrule(lr){5-7} \cmidrule(lr){8-10} \cmidrule(lr){11-13} 
		& P & R & F1 & P & R & F1 & P & R & F1 & P & R & F1 \\
		\hline \addlinespace
		Majority Baseline & 0.00 & 0.00 & 0.00 & 0.00 & 0.00 & 0.00 & 0.51 & 1.00 & 0.67 & 0.26 & 0.51 & 0.34 \\
		%		& & & & & & & & & & & & & & &\\
		Trained with Target \\
		~~~~~~~~+ Silver + Related Task\\
		~~~~~~~~~~~~Mean &0.56&	0.54&	0.55&	0.48&	0.46&	0.47&	0.67&	0.71&	0.70&	0.60&	0.60&	0.60\\
		~~~~~~~~~~~~(SD) & 0.04	& 0.05&	0.01&	0.01&	0.05&	0.03&	0.01&	0.04&	0.01&	0.00&	0.01&	0.01\\
		Trained with Parent\_Target\\
		~~~~~~~~+ Silver + Related Task\\
		~~~~~~~~~~~~Mean&0.55&	0.6&	0.59&	0.51&	0.49&	0.50&	0.72&	0.72&	0.72&	0.64&	0.64&	0.63\\
		~~~~~~~~~~~~(SD) & 0.03&	0.04&	0.01&	0.04&	0.05&	0.02&	0.02&	0.06&	0.02&	0.00&	0.01&	0.01 \\
		\bottomrule
		
	\end{tabular}
	\caption{Detailed results (P, R, and F) predicting whether the \emph{Target} is Hate, Neutral or Counter-hate when the input is only the Target or the Parent\_Target. The results are using both Silver and pretraining on related tasks. We experiment with multiple runs using different random seeds and report the mean scores and their standard deviation. }
	\label{t:multipleruns-results}
\end{table*}
% End: Table 

% Start: Table 
\begin{table*}[ht!]
	\centering
	\small
	\begin{tabular}{lcccc}
		\toprule
		& Epochs & Batch size & Learning rate & Dropout \\
		\midrule
		Target &  5 & 16 & 1e-5 & 0.5\\
		~~~~+ Silver & 2 & 16 &  1e-5 & 0.5\\
		~~~~+ Related task & 2 & 8 &  1e-5 & 0.5 \\
		~~~~+ Silver + Related task & 4 & 16 &  1e-5 & 0.5  \\
		\bottomrule
		
	\end{tabular}
	\caption{Hyperparameters used to fine-tune RoBERTa individually for each training setting. We accept default settings for the other hyperparameters as defined in the implementation by \citet{phang2020jiant}. }
	\label{t:hyperparameters}
\end{table*}
% End: Table 

\section{Detailed Results}
\label{sec:appendix}
Table \ref{t:detailed-results} presents detailed results complementing Table \ref{t:model-results} in the paper. 
We provide Precision, Recall and weighted F1-score using each related task for pretraining when the input is Target and Parent\_Target respectively in Table \ref{t:detailed-results}.

 Table \ref{t:multipleruns-results} presents the mean scores of Precision, Recall and weighted F1-score and their standard deviation when we use both Silver data and pretraining on related tasks with different random seeds. 
 The results are consistent with the findings in our study: adding the \textit{Parent} does improve the performance compared to the system that does not (0.63 vs. 0.60).

\section{Hyperparamters to Fine-tune the Systems}
\label{sec:appendix}

The neural model takes about half an hour on average to train on a single GPU of NVIDIA TITAN Xp. 
We use an implementation by \citet{phang2020jiant} and fine-tune the RoBERTa (base architecture; 12 layers) \cite{DBLP:journals/corr/abs-1907-11692} model for each of the four training settings. 
For each setting, we set the hyperparameters to be the same when the textual input is Target and Parent\_Target respectively. 
Hence we only report tuned hyperparameters for each setting when the input is Target in Table \ref{t:hyperparameters}.

\end{document}